\pdfoutput=1
\documentclass[11pt]{article}

\usepackage[final]{acl}

\usepackage{times}
\usepackage{tikz}  % TikZ 的基础包

\usepackage{siunitx}  
\usepackage{latexsym}
\usepackage{hyperref}
\usepackage{url}
\usepackage{amsthm}  % Add this line
\usepackage{amssymb}

\usepackage{wrapfig}
\usepackage{graphicx}
\usepackage{booktabs} 
\usepackage{multirow} 
\usepackage{array}  
\usepackage{amsmath}
\usepackage{algorithm}
\usepackage{algpseudocode}
\usepackage{adjustbox}
\usepackage{enumitem}

\usepackage{listings} % For formatting code or prompts
\usepackage{xcolor}   % For color formatting
\usepackage{fancyvrb} % For fancier verbatim environments

\usepackage{float}
\usepackage{caption}
\usepackage{subcaption}
\theoremstyle{definition}

\usepackage{pgfplots}
\pgfplotsset{compat=1.18}

\usepackage[T1]{fontenc}
\usepackage[utf8]{inputenc}

\usepackage{microtype}
\usepackage{amsmath,amsthm,mathtools}

\usepackage{inconsolata}
\usepackage{multicol}
\usepackage{graphicx}

\usepackage{tcolorbox} % 用于创建彩色文本框
\usepackage{soulutf8}  % 支持UTF8编码的高亮命令
\usepackage{soul}      % 支持高亮命令
\tcbuselibrary{breakable}  
\definecolor{mylightblue}{RGB}{173,216,230} 
\definecolor{applegreen}{rgb}{0.565,0.933,0.565}
\definecolor{mylightsalmon}{rgb}{1.0,0.627,0.478}

\newcommand{\hlgreen}[1]{{\sethlcolor{applegreen}\hl{#1}}}
\newcommand{\hlsalmon}[1]{{\sethlcolor{mylightsalmon}\hl{#1}}}
\title{SearchRAG: Can Search Engines Be Helpful for LLM-based\\ Medical Question Answering?}

\author{
 \textbf{Yucheng Shi\textsuperscript{1}},
 \textbf{Tianze Yang\textsuperscript{1}},
 \textbf{Canyu Chen\textsuperscript{2}},
\\
 \textbf{Quanzheng Li\textsuperscript{3}},
 \textbf{Tianming Liu\textsuperscript{1}},
 \textbf{Xiang Li\textsuperscript{3}},
 \textbf{Ninghao Liu\textsuperscript{1}},
\\
\\
 \textsuperscript{1}University of Georgia,
 \textsuperscript{2}Illinois Institute of Technology,
\\
 \textsuperscript{3}Massachusetts General Hospital and Harvard Medical School,
}

\begin{document}
\maketitle
\begin{abstract}
Large Language Models (LLMs) have shown remarkable capabilities in general domains but often struggle with tasks requiring specialized knowledge. Conventional Retrieval-Augmented Generation (RAG) techniques typically retrieve external information from static knowledge bases, which can be outdated or incomplete, missing fine-grained clinical details essential for accurate medical question answering. In this work, we propose SearchRAG, a novel framework that overcomes these limitations by leveraging real-time search engines. Our method employs synthetic query generation to convert complex medical questions into search-engine-friendly queries and utilizes uncertainty-based knowledge selection to filter and incorporate the most relevant and informative medical knowledge into the LLM's input. Experimental results demonstrate that our method significantly improves response accuracy in medical question answering tasks, particularly for complex questions requiring detailed and up-to-date knowledge.
\end{abstract}

\section{Introduction}
\begin{figure}[ht]
    \centering
    \includegraphics[width=0.9\linewidth]{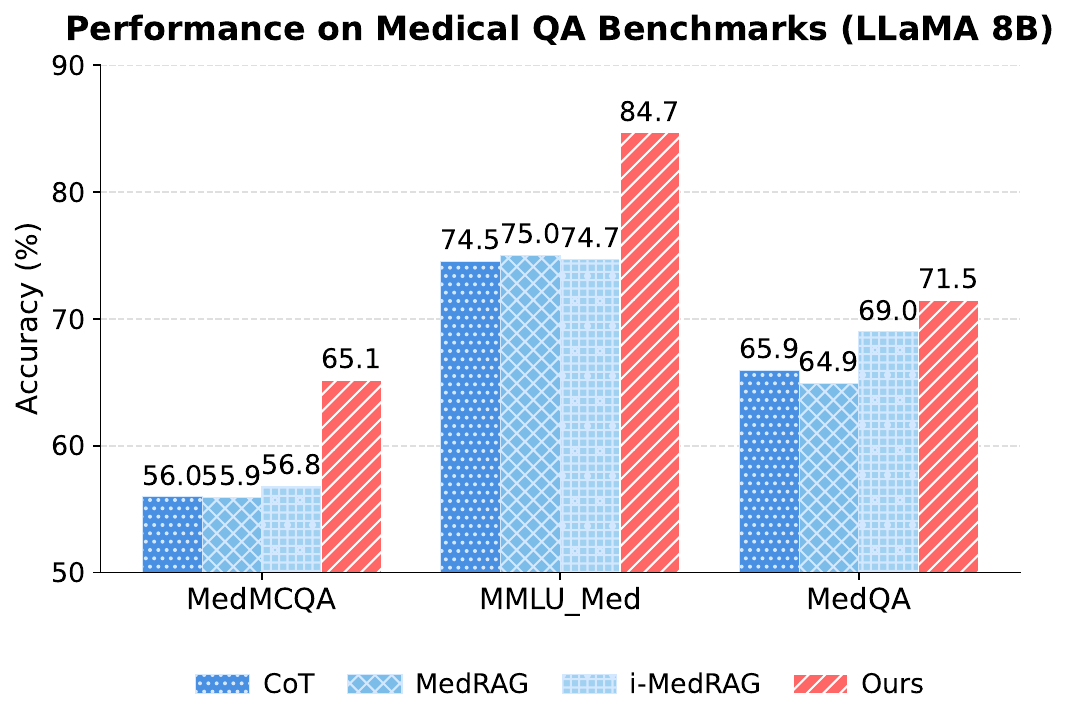}
    \caption{Performance comparison of different methods on medical QA benchmarks using LLaMA 8B. Our proposed method consistently outperforms baseline approaches (CoT), conventional RAG (MedRAG), and iterative RAG (i-MedRAG) across all three datasets (MedMCQA, MMLU\_Med, and MedQA), demonstrating significant improvements in accuracy.}
    \label{fig:performance_comparison}
    \vspace{-3pt}
\end{figure}

Large language models (LLMs) have shown proficiency in handling general-domain question answering tasks~\cite{brown2020language}.
However, they often face challenges when dealing with specialized fields that require up-to-date or domain-specific knowledge, such as medical question answering~\cite{jin2021disease, pmlr-v174-pal22a}. 
The limitation arises because the LLM parameters remain fixed after training, preventing the model from incorporating new medical knowledge essential for precise decision-making.

Retrieval-Augmented Generation (RAG) methods have been proposed to address this issue by incorporating external knowledge sources into the LLMs' input, thereby enhancing performance with additional context~\cite{guu2020retrieval, wang2021kepler}. 
Traditional RAG approaches typically retrieve information from static knowledge bases or knowledge graphs~\cite{shi2023mededit, xiong2024benchmarking}. 
However, these approaches face significant limitations. Recent studies~\cite{xiong2024benchmarking} show that RAG performance is heavily dependent on the quality and coverage of the knowledge corpus used. 
In many instances, using RAG actually reduces accuracy because the used knowledge bases is outdated or incomplete, missing fine-grained details critical for accurate medical decision-making~\cite{xiong2024benchmarking}.

In this work, we propose to tackle these challenges by utilizing search engines to enhance RAG systems. 
Search engines play a vital role for medical professionals, who rely on them to access clinical and medical knowledge that supports informed medical decision-making~\cite{lykke2012doctors}. 
By leveraging search engines, we can equip LLMs with real-time access to peer-reviewed research, clinical trial data, and up-to-date treatment protocols that are essential for effective medical decision support~\cite{yoran2023making}.

However, directly using the original medical questions or LLM-written queries as search engine input often yields suboptimal results. 
This is because original medical questions typically contain redundant contextual information that can mislead the search focus.
Moreover, current LLMs are not trained to align with search engines; therefore, they lack the inherent capability to generate queries optimized for search engines, which require specific structures and operators like Boolean logic and MeSH terms to function effectively~\cite{google_search_quotes_2022, ahrefs2024}. 
This misalignment between the LLM-written queries and the requirements of search engines leads to the retrieval of irrelevant or low-quality information, ultimately degrading the RAG performance~\cite{cuconasu2024power}.
This raises a critical research question: %\textbf{how to exploit search engines for LLM-based Medical QA?}
\textbf{How to help current LLMs with search engines for effective medical knowledge retrieval?}

To address this challenge, we propose \textbf{SearchRAG}, a novel retrieval-augmented generation framework that 
enhances the alignment between LLMs and search engines in inference-time through \emph{synthetic query generation} and \emph{uncertainty-based query selection}. 
Our approach first utilizes an LLM to generate a large and diverse set of search queries by repeatedly sampling with high temperature, based on the original medical question. 
These synthetic queries are designed to capture different aspects of the medical question in a search-engine friendly form. 
While not all generated queries are optimal, the diversity and scale of queries ensure that some candidates will effectively capture the clinical intent.
Then, to identify the most effective queries, our framework employs an uncertainty-based selection mechanism that evaluates the knowledge snippets retrieved by each query. 
By measuring the LLM’s uncertainty reduction when incorporating different knowledge snippets, we retain only those that contribute the most to improving model confidence.  

Through extensive experiments on medical QA benchmarks (Figure~\ref{fig:performance_comparison}), we demonstrate three key contributions:

\begin{itemize}[leftmargin=*]
    \item \textbf{Search Engine-Aligned Query Generation}: We propose SearchRAG, a novel method that bridges the gap between LLM capabilities and search engine requirements through synthetic query generation, specifically optimized for medical information retrieval.
    \item \textbf{Uncertainty-Based Query Selection}: We develop a selection mechanism that uses the LLM's internal entropy to identify and prioritize the most clinically relevant information from multiple retrieval candidates.  
    \item \textbf{Effective Performance Improvement}: We empirically show that our framework improves the LLaMA 8B model by an average 12.61\% compared to baseline methods in answer accuracy in medical QA tasks.
\end{itemize}

The paper is structured as follows: Section~\ref{sec:methodology} details our dual-component architecture for query generation and knowledge selection, Section~\ref{sec:experiments} presents comparative evaluations across multiple medical benchmarks, Section~\ref{related_work} discusses related work, and Section~\ref{sec:conclusion} concludes our paper.

\section{Methodology}
\label{sec:methodology}
\begin{figure*}
    \centering
    \includegraphics[width=1\linewidth]{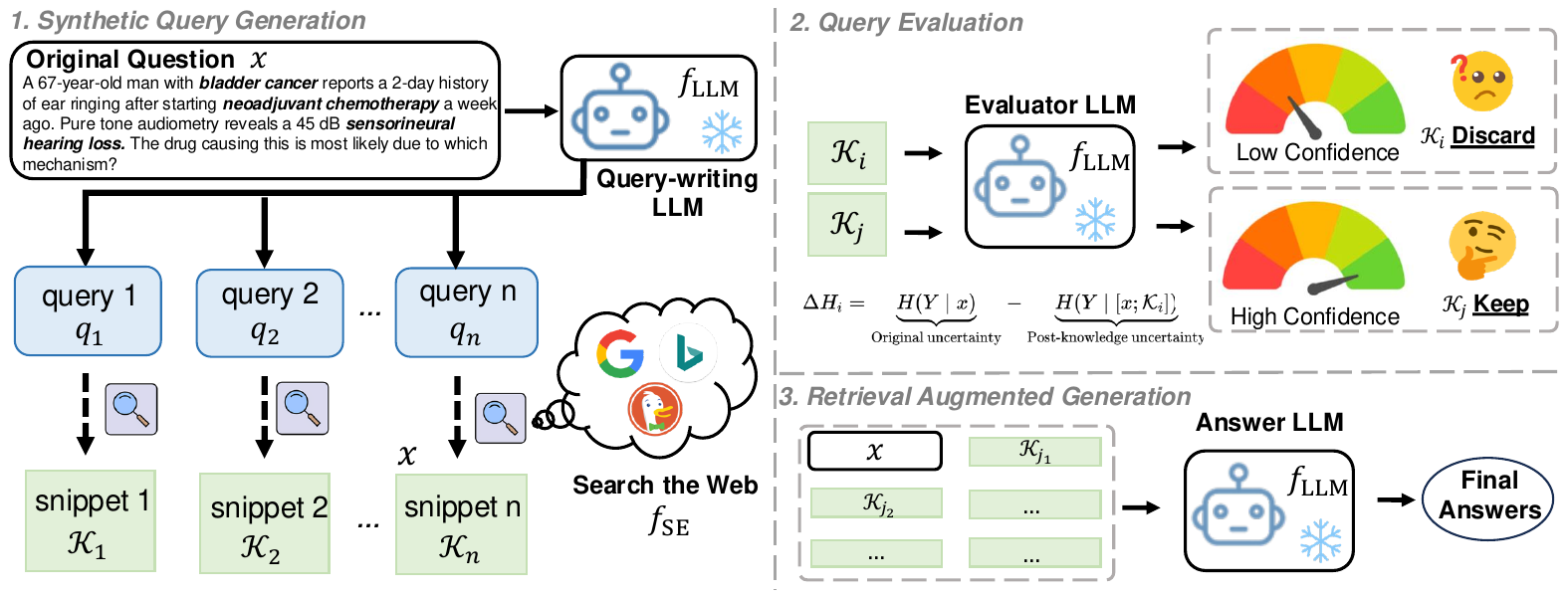}
    \caption{Overview of SearchRAG: Our framework first transforms complex medical questions into search-optimized synthetic queries through repeated sampling. The retrieved knowledge snippets are then filtered using uncertainty-based selection to identify the most relevant information for enhancing LLM responses.}
    \label{fig:enter-label}
\end{figure*}

In this section, we introduce SearchRAG, an innovative RAG framework aimed at enhancing LLMs' ability 
to effectively use search engines for knowledge retrieval, especially when addressing complex and specialized medical questions.

\subsection{Problem Formulation and Overview}

Retrieval-Augmented Generation (RAG) integrates external knowledge into the input of LLMs to enhance their responses \cite{guu2020retrieval, wang2021kepler}. Formally, given an input prompt $x$, RAG retrieves a set of knowledge snippets $\mathcal{K} = \{ \mathcal{K}_1, \mathcal{K}_2, \dots, \mathcal{K}_n \}$ using a retrieval function $f_{\text{ret}}$:
$
\mathcal{K} = f_{\text{ret}}(x),
$
and generates a response $y$ using model $f_{\text{LLM}}$ conditioned on both the input and the retrieved knowledge:
\begin{equation}
y = f_{\text{LLM}}(x, \mathcal{K}).
\end{equation}
However, when dealing with complex and verbose medical questions, such as the one shown in Figure~\ref{fig:enter-label}, directly using $x$ for retrieval can lead to suboptimal results. 
This is because such questions often include extraneous details that can mislead search engines and result in the retrieval of irrelevant information~\cite{jin2020disease, shi2023mededit, xiong2024benchmarking}.
To overcome these limitations, our proposed SearchRAG modifies the RAG framework by introducing two key components:
\begin{itemize}[leftmargin=*]
    \item A \emph{Synthetic Query Generation Module} that generates candidate queries by transforming original questions into search-engine-friendly form. %through repeated high-temperature sampling.
    \item An \emph{Uncertainty-Based Query Selection} mechanism that leverages the LLM's internal uncertainty to select the most informative knowledge snippets.
\end{itemize}
By integrating these components, we enhance the retrieval of pertinent information to answer questions, ultimately improving response accuracy and relevance.

\subsection{Synthetic Query Generation Module}

Medical questions are often too complex to be directly used as search engine queries.
To address this challenge, SearchRAG proposes to use synthetic queries generated by LLMs to improve the retrieval of knowledge pertinent to answering the original question.
The need for synthetic queries stems from two fundamental challenges in medical question answering:

\begin{itemize}[leftmargin=*]
    \item \textbf{Complexity of Medical Questions}: Medical questions often contain specialized terminology and multi-hop clinical logic~\cite{jin2020disease}.
    A single condition may be referred to by different medical terms (e.g., "myocardial infarction" and "heart attack"), making it challenging to match relevant content comprehensively.
    Additionally, critical retrieval keywords may not appear explicitly in the original question, which hinders effective retrieval.
    \item \textbf{Constraints of Search Engine}: Existing search engines employ algorithms and ranking systems that favor well-structured, concise queries with clear intent~\cite{google_search_quotes_2022}.  
    Medical questions usually fail to meet these requirements, thus preventing the retrieval of relevant medical content.
\end{itemize}

Our solution leverages the same LLM we aim to enhance through RAG to generate synthetic queries, eliminating the need for additional models.
We design a specialized prompt template $p$ (detailed in the Appendix) that guides the LLM to decompose and transform the original medical question into search queries. More specifically, the LLM first analyzes the medical question to extract key clinical entities, reformulates them using standardized medical terminology, and finally structures the queries with search-friendly syntax.
Given an original prompt $x$ and our template $p$, we generate a set of synthetic queries through repeated sampling:
\begin{equation}
\{ q_1, q_2, \dots, q_m \} = \bigcup_{i=1}^m f_{\text{LLM}}(x, p).
\end{equation}
Here, we employ a high-temperature sampling strategy to generate a \emph{large and diverse} set of candidate queries $\{ q_1, q_2, \dots, q_m \}$.
Example queries for a question $x$ are shown in Figure~\ref{fig:enter-label} with \{$q_1$:\emph{``ototoxic effects of platinum-based drugs"}; $q_2$: \emph{``mechanisms cisplatin hearing loss"}; $q_3$: \emph{``Proteasome inhibitor drug effects on inner ear"}\}.
These queries contain concise keywords suitable for search engines and cover multiple aspects of question $x$.
Among them, $q_2$ is the most informative one\footnote{A Google search for $q_2$ returns: \emph{``...cisplatin is retained for months to years. It can cause DNA damage, inhibit protein synthesis..."}, which corresponds to the correct answer:\emph{``Cross-linking of DNA"}. Though not mentioned in the original question, \emph{cisplatin} is a well-known chemotherapy drug.}. 
 We provide more details in Section~\ref{case_study} and more query examples in Appendix.

Our strategy is inspired by inference-time scaling law~\cite{brown2024large}: while not every query can retrieve optimal information, some candidates will be more effective than others at retrieving useful knowledge. The expanded candidate pool increases the probability of obtaining high-quality knowledge snippets.
To fully leverage this variability, our subsequent uncertainty-based selection mechanism (detailed in the next section) will identify and utilize the most effective ones.

\subsection{Align Search Engines to LLMs via Uncertainty-Based Query Selection}

While we generate a large number of synthetic queries, identifying the most effective ones is a non-trivial task.  
To address this, we propose to
leverage the LLM's internal uncertainty as a reward signal to guide query selection.
Specifically, we assess the uncertainty reduction associated with each retrieved snippet, treating it as a measure of information gain~\cite{shi2024retrieval}. Queries that retrieve knowledge leading to greater uncertainty reduction are prioritized, ensuring that only the most informative snippets contribute to RAG.  

Our approach aims to align search engine to LLMs by helping LLMs effectively use web search to retrieve knowledge to answer given questions.
By using uncertainty as a reward, our system selects queries that maximize knowledge utility, refining the alignment process on a per-query basis.  
Unlike tuning-based alignment~\cite{ma2023query}, our method optimizes in inference-time without modifying model parameters, making it robust to diverse retrieval sources and evolving topics.

For each query $q_i$, our search engine function $f_{\text{SE}}$ retrieves structured knowledge snippets:
\begin{equation}
\mathcal{K}_i = f_{\text{SE}}(q_i), \quad i = 1, 2, \dots, m.
\end{equation}
The retrieved snippets are concise extracts from search engine results, including titles and short descriptions for each webpage. We provide a case study of the search engine results in the Appendix. 
Each retrieved snippet augments the original prompt through concatenation:
$x_i' = [x; \mathcal{K}_i].$

We estimate the LLM's uncertainty in generating responses based on these augmented inputs using Shannon entropy~\cite{cover1999elements}.
Let $Y$ be the random variable representing possible responses $y$ generated by LLMs. 
Since computing the full response distribution is infeasible~\cite{shi2024retrieval,cover1999elements}, we approximate uncertainty using the entropy of the first token, defined as:  
\begin{equation}
H(Y \mid X) = -\sum_{w \in \mathcal{V}} p(w \mid X) \log_2 p(w \mid X),
\end{equation}
where $p(w \mid X)$ is the probability of generating token $w$ as the first token across the vocabulary $\mathcal{V}$. 
This approximation is suitable because medical questions are formulated as multiple-choice questions in our setting. 
We prompt the model to directly output option labels (\texttt{ABCD}) as the first token. 
Thus, the first token entropy directly reflects the model's confidence in different answer choices, making this uncertainty approximation both computationally efficient and semantically meaningful for our task.
For each knowledge snippet, we compute the reduction in conditional entropy:
\begin{equation}
\Delta H_i = \underbrace{H(Y\mid x)}_{\text{Original uncertainty}} - \underbrace{H(Y\mid x_i')}_{\text{Post-retrieval uncertainty}}.
\end{equation}
A positive value of $\Delta H_i$ indicates that the knowledge snippet $\mathcal{K}_i$ has reduced the model uncertainty. 
The most informative snippets are aggregated into a final knowledge set:
\begin{equation}
\mathcal{K}^* = \bigcup \left\{ \mathcal{K}_i \mid \Delta H_i > 0 \right\}.
\end{equation}
This inference-time alignment leverages the model internal states to ensure that only the knowledge snippets capable of increasing prediction certainty are used for the final generation.

\begin{table*}[htp]
    \centering
    \caption{Comparison of knowledge source characteristics.}
    \label{tab:source_comparison}
    \begin{tabular*}{\textwidth}{@{\extracolsep{\fill}}lcccc@{}}
    \toprule
    Source & \small Data Volume & \small Source Diversity &  \small External Space Use & \small External Memory Use \\
    \midrule
    Textbooks~\cite{jin2021disease} & $\times$ & $\times$ & $\triangle$ & $\triangle$\\
    PubMed~\cite{pile} & $\triangle$ & $\triangle$ & $\times$ & $\times$ \\
    Search Engine~\cite{GoogleSearch} & $\checkmark$ & $\checkmark$ & $\checkmark$ & $\checkmark$ \\
    \bottomrule
    \end{tabular*}
    \smallskip
    \parbox{\textwidth}{\footnotesize $\checkmark$: Strong capability \hspace{1em} $\triangle$: Moderate capability \hspace{1em} $\times$: Weak capability}
\end{table*}

\subsection{Integration into the RAG Framework}
The complete integration of SearchRAG's components into the RAG framework operates as Algorithm~\ref{algo}.
The framework begins by generating diverse search queries through repeated sampling from the LLM (line 3). 
Next, each query is input into a search engine to retrieve knowledge (line 5). 
The system then evaluates the retrieved content by measuring how effectively each snippet reduces the model's prediction uncertainty (line 6).
Finally, confidence-filtered knowledge is aggregated for the final answer generation (lines 8, 9), 
ensuring that only clinically relevant information is utilized to assist the diagnosis.
\begin{algorithm}[H]
    \caption{SearchRAG}
    \begin{algorithmic}[1]
        \State \textbf{Input}: Original question $x$
        \State \textbf{Output}: Response $y$
        \State $\{q_1,\ldots,q_m\} \gets f_{\text{LLM}}(x, p)$ \Comment{Synthetic query generation}
        \For{$i \in \{1,\ldots,m\}$}
            \State $\mathcal{K}_i \gets f_{\text{SE}}(q_i)$ \Comment{Search engine retrieval}
            \State $\Delta H_i \gets H(Y|X=x) - H(Y|X=[x;\mathcal{K}_i])$ \Comment{Uncertainty reduction}
        \EndFor
        \State $\mathcal{K}^* \gets \bigcup \left\{ \mathcal{K}_i \mid \Delta H_i > 0 \right\}$ \Comment{Knowledge selection}
        \State $y \gets f_{\text{LLM}}(x, \mathcal{K}^*)$
        \State \Return $y$
    \end{algorithmic}
    \label{algo}
\end{algorithm}

\section{Experiments}
\label{sec:experiments}
Our experimental evaluation addresses three key research questions (RQs): 
RQ1: How does SearchRAG perform compared to existing methods on medical QA tasks? 
RQ2: How effective is our uncertainty-based knowledge selection strategy in improving performance? 
RQ3: How does scaling the number of synthetic queries affect the model's performance?
We evaluate through comparative benchmarks on medical datasets, focusing on quantitative performance metrics and qualitative analysis of retrieved knowledge.

\subsection{Experimental Settings}

We evaluate our approach against four baseline methods: one non-RAG method and three RAG-based methods.
The non-RAG baseline is Chain-of-Thought (CoT) prompting~\cite{wei2022chain}.
For RAG-based methods, we compare against: MedRAG using textbooks as the knowledge source with MedCPT retriever,
MedRAG using PubMed as the knowledge source with MedCPT retriever,
and i-MedRAG using textbooks as the knowledge source with MedCPT retriever~\cite{xiong2024benchmarking, xiong2024improving}.
A comparison of the different knowledge sources is shown in Table~\ref{tab:source_comparison}.
Additional implementation details, including hyperparameter choices, can be found in the Appendix.

We evaluate these methods on three datasets designed for medical question answering: 
MedQA~\cite{jin2021disease}, MMLU\_Med~\cite{hendrycks2020measuring}, and MedMCQA~\cite{pmlr-v174-pal22a}. 
MedQA consists of real-world medical licensing exam questions, 
MMLU\_Med is a specialized subset of a larger multi-domain challenge focusing on medical topics, 
and MedMCQA contains multiple-choice medical questions that require domain-specific reasoning. 
We follow the evaluation protocols established by MedRAG and i-MedRAG~\cite{xiong2024benchmarking, xiong2024improving}.

For the LLMs, we experiment with two instruction-tuned variants of LLaMA 3.1, specifically the 8B and 70B parameter versions~\cite{dubey2024llama}. 
We utilize Google Search\footnote{Third-party search API provider. Retrieved search snippets are used as relevant knowledge for RAG augmentation. \url{https://serper.dev/}} as our search engine provider.
For the 70B model, we employ INT4 quantization to reduce memory requirements while maintaining performance. 
All variants are utilized without additional fine-tuning for any baseline approach. 
All inference procedures are carried out on A6000 GPUs. 
Additional experiment details will be provided in the Appendix.
\begin{table*}[htp]
    \centering
    \caption{Performance comparison of different methods on medical QA datasets.}
    \begin{tabular}{lcccccc}
        \toprule
        & \multicolumn{2}{c}{\textbf{MedMCQA}} & \multicolumn{2}{c}{\textbf{MMLU\_Med}} & \multicolumn{2}{c}{\textbf{MedQA}} \\
        \cmidrule(lr){2-3} \cmidrule(lr){4-5} \cmidrule(lr){6-7}
        \textbf{Method} & Accuracy& Improve& Accuracy& Improve& Accuracy& Improve\\
        \midrule
        \multicolumn{7}{l}{\textbf{LLaMA 3.1-8B}} \\
        CoT &  55.96& - & 74.52& - & 65.91& - \\
        MedRAG (Textbooks) &  55.89&  -0.13\% &  74.98&  0.62\% & 64.89&  -1.55\% \\
        MedRAG (PubMed) &  50.87&  -9.11\% &  71.28&  -4.32\% & 60.41&  -8.48\% \\
        i-MedRAG &  56.80&  +1.65\%& 74.70& +0.25\%& 68.97& +5.07\%\\
        \textbf{Our RAG} &   \textbf{65.14}&   \textbf{+16.16\%}& \textbf{84.67}&  \textbf{+13.59\%}& \textbf{71.49}&\textbf{+8.09}\%  \\
        \midrule
        \multicolumn{7}{l}{\textbf{LLaMA 3.1-70B}} \\
        CoT &  69.33& - &  87.53& - & 78.08& - \\
        MedRAG (Textbooks) &  69.23&  -0.14\% &  86.33&  -1.37\% & 79.34&  +1.61\% \\
        MedRAG (PubMed) &  69.95&  +0.90\% &  86.80&  -0.85\% & 79.26&  +1.49\% \\
        i-MedRAG &  69.23&  -0.14\% &  87.17&  -0.41\% & 79.58&  +1.89\% \\
        \textbf{Our RAG} &  \textbf{74.40}  & \textbf{+7.32\%} &  \textbf{90.95}&   \textbf{+3.92\%}&   \textbf{83.34}&   \textbf{+6.61\%} \\
        \bottomrule
    \end{tabular}
    \label{tab:main_results}
\end{table*}
\subsection{RQ1: Main Results}
Table~\ref{tab:main_results} demonstrates the effectiveness of SearchRAG across model scales and medical QA datasets. We have four findings:

First, conventional RAG approaches exhibit critical sensitivity to knowledge source selection, particularly in smaller models. 
Our experiments reveal that PubMed-based retrieval degrades 8B model performance by 9.11\% on MedMCQA compared to chain-of-thought baselines, while textbook-based retrieval provides only marginal gains (+0.62\% on MMLU\_Med). 
This validates our core hypothesis that static knowledge bases struggle to provide reliable augmentation, as noted in our analysis of search engine alignment challenges (Section 2.1).

Second, our dual-component architecture enables \textbf{consistent and substantial performance improvements} across all datasets and model scales. 
The 8B model with SearchRAG achieves absolute accuracy gains of +16.16\% on MedMCQA, +13.59\% on MMLU\_Med, and +8.09\% on MedQA compared to chain-of-thought baselines, significantly outperforming i-MedRAG's maximum gain of +5.07\% on MedQA. 
This performance advantage stems from our method's unique combination of search-optimized query generation and uncertainty filtering. 
Moreover, our synthetic queries enable better handling of complex clinical reasoning, particularly evident in MedMCQA where questions require synthesizing multiple medical concepts.

Third, the uncertainty-based filtering mechanism demonstrates model-agnostic effectiveness. 
Despite the 70B model's stronger baseline performance (69.33\% MedMCQA vs 55.96\% for 8B), SearchRAG still delivers substantial gains of +7.32\% on MedMCQA and +6.61\% on MedQA. 
This confirms our theoretical framework's scalability, rather than being subsumed by model capacity increases, our knowledge selection mechanism can continually provide complementary knowledge.

Finally, our method shows relatively modest gains with the smaller model on MedQA (+8.09\% for 8B), 
which often demands more intricate clinical reasoning. 
This discrepancy indicates that our method is particularly effective in situations where the primary limitation is the need for precise factual knowledge, 
rather than the ability to navigate complex reasoning chains.

In summary, these results empirically validate SearchRAG's two-stage alignment process. The significant improvements demonstrate effective search engine query optimization, while the consistent gains across model scales confirm the uncertainty filtering's generalizability.

\subsection{Ablation Studies}

\subsubsection{RQ2: Effectiveness of Knowledge Selection}
To validate our uncertainty-based knowledge selection mechanism, we conduct experiments on 350 randomly sampled instances from each dataset using both model variants. Contrary to the intuition that unfiltered knowledge provides richer context, Table \ref{tab:filtering_ablation} reveals that selective filtering consistently improves performance, particularly for smaller models.
\begin{table*}[h]
    \centering
    \caption{Performance comparison with/without knowledge filtering across model scales and datasets.}
    \begin{tabular*}{\textwidth}{@{\extracolsep{\fill}}lcccccc@{}}
        \toprule
        & \multicolumn{2}{c}{\textbf{MedMCQA}} & \multicolumn{2}{c}{\textbf{MMLU\_Med}} & \multicolumn{2}{c}{\textbf{MedQA}} \\
        \cmidrule(lr){2-3} \cmidrule(lr){4-5} \cmidrule(lr){6-7}
        \textbf{Model} & \multicolumn{1}{l}{Unfiltered} & \multicolumn{1}{l}{Filtered} & \multicolumn{1}{l}{Unfiltered} & \multicolumn{1}{l}{Filtered} & \multicolumn{1}{l}{Unfiltered} & \multicolumn{1}{l}{Filtered} \\ 
        \midrule
        %\textbf{8B Model} \\
        \textbf{8B} & 60.57 & \textbf{64.86} (+7.08\%)& 81.14 & \textbf{84.86} (+4.58\%)& 67.71& \textbf{72.29} (+6.76\%)\\
        \addlinespace[0.5em]
        
        %\textbf{70B Model} \\
        \textbf{70B} & 70.86 & \textbf{74.57} (+5.24\%)& 88.29 & \textbf{89.71} (+1.61\%)& 82.57 & \textbf{83.43} (+1.04\%)\\
        \bottomrule
    \end{tabular*}
    \label{tab:filtering_ablation}
    \vspace{-5pt}
\end{table*}

There are two key insights as follows: (1) The LLaMA 8B model shows 4-7\% absolute gains across all datasets, indicating smaller models' vulnerability to irrelevant information: unfiltered knowledge introduces distracting noise that misleads reasoning. 
(2) While the LLaMA 70B model exhibits more robustness, it still benefits from filtering, suggesting even powerful models cannot fully compensate for low-quality context. 
This demonstrates the effectiveness of our uncertainty-based selection mechanism: By selectively presenting only the most relevant information, we help models avoid distraction from excessive content while maintaining useful knowledge signals.
%Our filtering mechanism acts as a cognitive focuser, helping models concentrate on truly pertinent knowledge.

\subsubsection{RQ3: Impact of the Number of Total Generated Queries}

\begin{figure*}[htbp]
    \centering
    \begin{tikzpicture}
        \begin{axis}[
            width=0.33\textwidth,
            height=4cm,
            title=MedMCQA,
            xlabel=Generated Queries,
            symbolic x coords={0, 4,16,32},
            xtick=data,
            ymin=55, ymax=68,
            grid=both,
            grid style={line width=.1pt, draw=gray!20},
            major grid style={line width=.2pt,draw=gray!40},
            bar width=10pt,
            nodes near coords={\pgfmathprintnumber[precision=1]\pgfplotspointmeta\%},
            every node near coord/.append style={font=\scriptsize, rotate=0, yshift={3pt}},
            ytick={55,60,65},
            yticklabel style={font=\footnotesize},
            xticklabel style={font=\footnotesize}
        ]
        \addplot+[ybar, fill=blue!50!white, draw=blue] coordinates {
            (0, 57.14)
            (4, 58.57)
            (16, 62.86)
            (32, 64.86)
        };
        \end{axis}
    \end{tikzpicture}
    \hspace{2pt}
    \begin{tikzpicture}
        \begin{axis}[
            width=0.33\textwidth,
            height=4cm,
            title=MMLU\_Med,
            xlabel=Generated Queries,
            symbolic x coords={0,4,16,32},
            xtick=data,
            ymin=74, ymax=88,
            grid=both,
            grid style={line width=.1pt, draw=gray!20},
            major grid style={line width=.2pt,draw=gray!40},
            bar width=10pt,
            nodes near coords={\pgfmathprintnumber[precision=1]\pgfplotspointmeta\%},
            every node near coord/.append style={font=\scriptsize, rotate=0, yshift={3pt}},
            ytick={74,78,82,86},
            yticklabel style={font=\footnotesize},
            xticklabel style={font=\footnotesize}
        ]
        \addplot+[ybar, fill=blue!50!white, draw=blue] coordinates {
            (0, 76.86)
            (4, 79.43)
            (16, 81.43)
            (32, 84.86)
        };
        \end{axis}
    \end{tikzpicture}
    \hspace{2pt}
    \begin{tikzpicture}
        \begin{axis}[
            width=0.33\textwidth,
            height=4cm,
            title=MedQA,
            ylabel=Accuracy (\%),
            xlabel=Generated Queries,
            symbolic x coords={0,4,16,32},
            xtick=data,
            ymin=60, ymax=76,
            grid=both,
            grid style={line width=.1pt, draw=gray!20},
            major grid style={line width=.2pt,draw=gray!40},
            bar width=10pt,
            nodes near coords={\pgfmathprintnumber[precision=1]\pgfplotspointmeta\%},
            every node near coord/.append style={font=\scriptsize, rotate=0, yshift={3pt}},
            ytick={60,65,70,75},
            yticklabel style={font=\footnotesize},
            xticklabel style={font=\footnotesize}
        ]
        \addplot+[ybar, fill=blue!50!white, draw=blue] coordinates {
            (0, 62.85)
            (4, 63.71)
            (16, 67.14)
            (32, 72.29)
        };
        \end{axis}
    \end{tikzpicture}
    \caption{Comparison of model accuracies for MedMCQA, MMLU\_Med, and MedQA across different numbers of generated queries. Note that the 0-query baseline directly uses the original question for retrieval. As the number of generated queries increases, the performance of RAG improves significantly.}
    \label{fig:results_overview}
    \end{figure*}
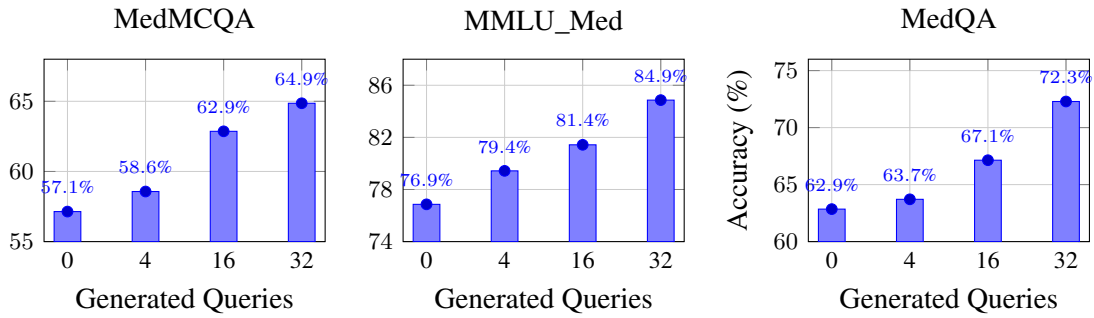

We conducted an evaluation to understand how the number of generated queries affects model performance, using 350 randomly sampled instances. 
The configurations tested included using 0 (the original question), 4, 16, and 32 generated queries. 
Figure~\ref{fig:results_overview} illustrates the impact of varying the number of queries on performance.

Our analysis reveals two key findings: 
First, using only the original questions (0 query) resulted in limited RAG effectiveness, with performance even degrading compared to the non-RAG baseline, achieving only 62.9\% on the MedQA. 
\textbf{This suggests unmodified medical questions fail to effectively leverage search engines for retrieving relevant information}, and the low performance confirms that these medical questions do not have direct matches in existing online knowledge sources.
Second, introducing synthetic query generation led to improvements across all datasets. Performance scaled with the number of generated queries, peaking with 32 queries. %: 64.9\% on MedMCQA, 84.9\% on MMLU\_Med, and 72.3\% on MedQA. 
This clear correlation between query count and accuracy indicates that a larger query set helps capture more diverse and relevant knowledge to enhance the RAG process.

\subsubsection{Case Study}
\label{case_study}
This case study demonstrates how SearchRAG handles a clinical question about oral contraceptive effects. The scenario involves a 17-year-old patient prescribed contraceptive pills, examining which condition's risk is decreased by this medication. Through high-temperature sampling, the model generates diverse search queries, retrieving evidence that oral contraceptives reduce ovarian cancer risk while increasing risks for cervical and breast cancers (See Facts [1][2][7]). While the system initially incorrectly suggested cervical cancer as having reduced risk, the retrieved facts enabled the model to correct its response to ovarian cancer. This case shows how SearchRAG's query generation and knowledge selection can transform an incorrect initial assessment into an accurate, evidence-based conclusion.
More cases are provided in Appendix.

\begin{tcolorbox}[breakable, title={Case Study 1: }, colback=white, colframe=black, colbacktitle=white, coltitle=black, fonttitle=\bfseries, rounded corners]
\small
\textbf{Question:}
A 17-year-old girl comes to the physician because of an 8-month history of severe acne vulgaris over her face, upper back, arms, and buttocks. Treatment with oral antibiotics and topical combination therapy with benzoyl peroxide and retinoid has not completely resolved her symptoms. Examination shows oily skin with numerous comedones, pustules, and scarring over the face and upper back. Long-term therapy is started with \textit{combined oral contraceptive pills}. This medication \textit{decreases the patient's risk} developing of which of the following conditions?
\vspace{5pt} %

\textbf{Choices:} 
[A.] Hypertension, [B.] Ovarian cancer, [C.] Cervical cancer, [D.] Breast cancer

\vspace{2pt} 

\textbf{Ground Truth: B (Ovarian cancer)}
\vspace{2pt} %

\textbf{Synthetic Queries:} 

[1] 'oral contraception use reducing ovarian cancer risk', [2] 'risk of hormonal cancers contraceptive pills', \textcolor{gray}{[3] 'Oral contraceptive pill preventive effects on cancer'}, \textcolor{gray}{[4] 'Carcinogenic effects of estrogen therapy'}, \textcolor{gray}{[5] 'oral contraceptive linked conditions breast cancer'}, \textcolor{gray}{[6] 'combined oral contraceptive risk breast cancer'}, [7] 'Oral contraceptive pills cancer prevention effects on patients', \textcolor{gray}{[8] 'Effect of combined oral contraceptives on cancer risk'}...
\vspace{2pt} %

\textbf{Filtered Facts:} 

[1] Ovarian cancer: Women who have ever used oral contraceptives \hlgreen{have a 30\% to 50\% lower risk of ovarian cancer} than women who have never used oral contraceptives.

[2] The longer a woman uses oral contraceptives, the greater the \hlgreen{increase in her risk of cervical cancer}. One study found a 10\% increased risk ....

[7] Combined oral contraceptives (OCs) have been associated with a \hlgreen{reduced risk of ovarian and endometrial cancers}, an increased risk of cervical cancer and liver cancers, and an \hlgreen{increased risk of breast cancer} among recent users.

\vspace{2pt}

\textbf{Answer Before Retrieval:} \\
The patient is a 17-year-old girl with severe acne vulgaris that ... Considering the patient's age and sex, the most relevant condition to consider is related to reproductive health. Among the options provided, \hlsalmon{\textit{cervical cancer}} is a well-known risk that can be reduced with the use of oral contraceptive pills...
\vspace{2pt} 

\textbf{Answer after Retrieval:} \\
To determine the correct answer, let's analyze the information provided in the documents. 

1. The patient is a 17-year-old girl with ...

2. We need to find out which condition ...

3. According to the documents, combined oral contraceptives (OCs) have been associated with a \hlgreen{reduced risk of ovarian cancer}.

4. The documents also mention that \hlgreen{the longer a woman uses oral contraceptives, the greater the increase in her risk of cervical cancer}, but this is not a reduction in risk.

5. Additionally, OCs have been associated with an \hlgreen{increased risk of breast cancer} among recent users, but this is not a reduction in risk.

6. Finally, the documents mention that OCs have been associated with a \hlgreen{reduced risk of ovarian and endometrial cancers}, but the question specifically asks about ovarian cancer.

Therefore, based on the information provided, the correct answer is related to \hlsalmon{\textit{ovarian cancer}}.
\footnotetext{\hlgreen{\;\;\;}: Retrieved facts utilized in the final answer.}
\footnotetext{\hlsalmon{\;\;\;}: The selected answer.}

\end{tcolorbox}

\section{Related Work}
\label{related_work}
While large language models excel at question answering, they struggle with specialized medical knowledge~\cite{shi2023mededit}.
Traditional RAG systems have been proposed to mitigate these limitations by integrating external knowledge sources into the LLMs' input, thereby enhancing their ability to generate accurate and contextually relevant responses \cite{brown2020language, guu2020retrieval}. Nonetheless, these early RAG approaches predominantly relied on retrieving information from structured databases or knowledge graphs, which can be insufficient for handling the complexity and nuanced requirements of specialized medical queries \cite{jin2020disease, xiong2024benchmarking}.

Recent advancements in RAG systems have focused on improving the accuracy and relevance of retrieved medical information. 
Xiong et al.~\cite{xiong2024improving} propose i-MedRAG, which uses iterative follow-up questions to help LLMs better understand complex medical queries through dynamic information seeking.
Das et al.~\cite{das2024two} introduce a two-layer RAG framework and use soical media data to address challenges in low-resource medical question answering scenarios. 

Recent work has also explored uncertainty estimation in RAG systems for various purposes~\cite{ozaki2024understanding, li2024uncertaintyrag}. 
Some approaches use uncertainty to determine when to retrieve external information~\cite{ding2024retrieve, liu2024ctrla}, while others focus on improving robustness by reducing semantic inconsistencies from random chunking~\cite{li2024uncertaintyrag} or enhancing decoding~\cite{qiu2024entropy}. 
Despite these advancements, existing RAG frameworks still struggle with retrieving fine-grained and up-to-date knowledge, particularly for complex medical questions. 
This is because traditional knowledge bases and graphs often lack the granularity and timeliness needed for specialized medical queries.
Our approach addresses these limitations by utilizing search engines for real-time knowledge retrieval. 
 
\section{Conclusion}
\label{sec:conclusion}
In this paper, we presented SearchRAG, a novel retrieval-augmented generation framework designed to enhance large language models' performance in complex medical question-answering tasks. 
Our approach integrates synthetic query generation and uncertainty-based knowledge selection to optimize the retrieval of relevant information from search engines. 
Through extensive experiments, we demonstrated that SearchRAG significantly improves the accuracy and coherence of the model's responses compared to traditional RAG approaches. 
These findings suggest that our framework can effectively address the limitations of conventional RAG systems.

\section{Limitations} % Required for ACL !!
\label{sec:discussion}
One potential limitation of our approach is the reliance on search engines for knowledge retrieval. 
The results returned by search engines may sometimes include incorrect or unreliable information sources. 
In practical applications, it is crucial to have domain experts review the retrieved content to ensure its accuracy. 
Alternatively, restricting the search to a curated list of trusted websites can help mitigate this issue and improve the reliability of the information used by the model.

\section{Ethical Impact}
This research utilizes the LLaMA foundation model~\cite{dubey2024llama}, operating within the scope of its academic licensing agreement. Our implementation strictly adheres to the academic-use provisions specified in the license, with all applications limited to scholarly research purposes. The study draws upon three medical domain datasets: MedQA~\cite{jin2021disease}, MMLU\_Med~\cite{hendrycks2020measuring}, and MedMCQA~\cite{pmlr-v174-pal22a}, each employed in accordance with their respective usage guidelines and data governance frameworks. We have conducted thorough reviews to ensure compliance with data protection protocols, confirming the absence of personally identifiable information such as patient names or unique identifiers. Furthermore, our data processing protocols have verified that the content is appropriate and free from inappropriate material, maintaining high standards of research ethics and data integrity. Additionally, we utilized ChatGPT~\cite{hurst2024gpt} to assist with grammatical refinements during the writing process.

\bibliography{custom}

\appendix
\label{sec:appendix}

\section{Experimental Details}
\label{sec:appendix-experimental-details}

This section provides additional details about our experimental settings, dataset sources, retriever configurations, and prompt design used throughout the paper.

\subsection{Model Configurations and Hyperparameters}
\label{sec:model-hparams}

\paragraph{Base LLMs.} We use two variants of the LLaMA 3.1 model~\cite{dubey2024llama}, at 8B and 70B parameter scales. Both variants are instruction-tuned but not further fine-tuned for our experiments. For the 8B model, we employ \emph{bfloat16} for inference.
For the 70B model, we employ INT4 quantization to reduce memory usage. 

\paragraph{Inference Setup.} 
All experiments run on A6000 GPUs (48GB memory). The average GPU hours for our SearchRAG is roughly 0.018 hours/per medical question.
Unless noted otherwise, we set the maximum generation length to 512 tokens, with a \emph{do\_sample=False}. 
For synthetic query generation, we increase the temperature up to 2.0 to promote query diversity. 
We generate up to 32 candidate queries for each input question; in ablation studies, we vary this number to assess its impact. 

\paragraph{Uncertainty Computation.}
In Section~3.3 of the main paper, we measure uncertainty by approximating the Shannon entropy at the first-token output distribution.  
We use the final layer's logits after seeing each query-augmented context.  
Higher entropy implies lower confidence.  
The difference in entropy ($\Delta H$) between the original context and the query-augmented context determines whether to keep each snippet.

\subsection{Datasets}
\label{sec:datasets}

\paragraph{MedQA.} We use the benchmark from \citet{jin2021disease}, which contains multiple-choice questions from US medical licensing exams. We follow the standard split of 1,273 questions for evaluation. Each question has four answer choices (A--D). 

\paragraph{MMLU\_Med.} We use the medical subset of the Massive Multitask Language Understanding (MMLU) benchmark \cite{hendrycks2020measuring}, consisting of six biomedical subject areas (anatomy, clinical knowledge, professional medicine, human genetics, college medicine, and college biology), for a total of 1,089 test questions. Each question also has four multiple-choice options.

\paragraph{MedMCQA.} We adopt the dataset introduced by \citet{pmlr-v174-pal22a}, which consists of 4,183 medical multiple-choice questions from real-world medical entrance exams. Each question has four options, and we follow the official evaluation split for testing.

\subsection{Knowledge Sources and Retriever Setup}
\label{sec:data-source}

\subsubsection{PubMed Subset}
PubMed is a comprehensive repository of biomedical and life sciences articles~\cite{lu2011pubmed}. In total, it indexes over 36 million articles. For our study, we follow~\citet{xiong2024benchmarking} and utilize a curated subset of \textit{around 23.9 million} entries where each entry has a valid title and abstract.

\subsubsection{Textbooks Collection}
We also incorporate a set of 18 widely-used medical textbooks that cover fundamental subjects relevant for medical board examinations~\cite{jin2020disease}. 
These textbooks offer a broad range of clinically relevant facts and explanations.

\subsection{Retriever and Search Engine Details}
\label{sec:retriever}

\paragraph{Retriever Implementation.}
For \textbf{PubMed} and \textbf{Textbooks}, we follow a standard dense retrieval approach. We encode the document chunks using a domain-adapted bi-encoder (e.g., \texttt{MedCPT} \cite{xiong2024benchmarking}) and retrieve the top-$k$ chunks based on cosine similarity with the query vector. We typically set $k=32$ in our comparisons.

\paragraph{Web Search Interface.}
For the web-based retrieval, we employ a public search API\footnote{We used a third-party provider for Google Search results, specifically \url{https://serper.dev/}}. 
For each search, we employ the retrieved knowledge graph or answer box (if returned) and the top 3 organic results (title and snippet) for each generated query. The returned snippets typically include a short paragraph (50--200 words) that we concatenate to form the context. 
It is worth noting that we never included any AI overview content from Google.
We omit any images, ads, or sections lacking textual descriptions. 
By design, we do not scrape full web pages, which mitigates potential copyright concerns. 
Instead, we rely on the concise snippets from search results to guide the LLM.
\begin{tcolorbox}[colback=white,colframe=black!75,title=Example Response for "Apple"]
    \small
    \textbf{Knowledge Graph:}
    \begin{itemize}
        \item Title: Apple
        \item Description: American multinational technology company...
        \item Attributes: Headquarters: Cupertino, CA; CEO: Tim Cook
    \end{itemize}
    
    \textbf{Organic Results:}
    \begin{enumerate}
        \item \textbf{Apple Official Site}\\ 
        "Discover the innovative world of Apple and shop everything iPhone, iPad..."
        \item \textbf{Apple Wikipedia}\\ 
        "Apple Inc. is an American multinational technology company specializing in..."
    \end{enumerate}
    \end{tcolorbox}

\subsection{Prompt Design}
\label{sec:prompt-design}

\paragraph{Synthetic Query Generation Prompt.}
We design a specialized template to elicit concise, search-optimized queries from the LLM. 
Below is a simplified illustration:

\begin{tcolorbox}[colback=white, colframe=black, boxrule=0.6pt]
\textbf{System Role}: 
``You are a medical expert. Generate focused search queries that will help determine the correct relationship between medical concepts. Your queries should:
\begin{itemize}
    \item Target the specific medical association being tested
    \item Find evidence linking concepts in the question and options  
    \item Help differentiate between answer choices
\end{itemize}

Given this medical question and its answer options, identify what specific general medical knowledge is needed to correctly answer the question. Generate one most relevant retrieval inquiry that is:
\begin{itemize}
    \item 3--8 words long
    \item Focused on key medical terms
    \item Formatted like search engine input
    \item Targeting specific associations rather than general information
\end{itemize}

Output your search query after 'Search\_query:' and think step by step.''

\textbf{User Role}: 
``Question: [\textit{Original Medical Question}] \\
Possible Answers: A) \ldots \quad B) \ldots \quad C) \ldots \quad D) \ldots \\ 
Please provide a single short query.''
\end{tcolorbox}
We sample multiple times (with higher temperature) to produce a range of candidate queries.

\begin{tcolorbox}[colback=white, colframe=black, boxrule=0.6pt]
\textbf{System Role}: 
``You are a medical expert. Please pick the most likely option among A--D directly.''

\textbf{User Role}: 
``Information: [\textit{Snippet Text}] \\
Question: [\textit{Same Medical Question}] \\
Answer Choices: A) \ldots \ B) \ldots \ C) \ldots \ D) \ldots \\
Answer:''
\end{tcolorbox}
\paragraph{Uncertainty-Based Filtering Prompt.}
To measure how much each snippet reduces the model's uncertainty, we provide a short prompt to the model asking it to \textit{directly choose} an answer (A--D) given the retrieved snippet. The system's output distribution for that first token is used to compute approximate entropy. This prompt is concise.

We record the model's probability distribution over tokens to approximate entropy. A lower resulting entropy (compared to the base question without the snippet) indicates that snippet was informative.

\paragraph{Final Answer Prompt.}
After selecting informative snippets, we concatenate them and present them to the model alongside the original question. The final prompt is:

\begin{tcolorbox}[colback=white, colframe=black, boxrule=0.6pt]
\textbf{System Role}: ``You are a helpful medical expert.''
\newline
\textbf{User Role}: 
\begin{quote}
\small
``Below are some relevant excerpts:
\{\texttt{[Concatenated Snippets]}\}

Here is the question:
\{\texttt{[Medical Question]}\}

Possible Answers:
A) \ldots \quad B) \ldots \quad C) \ldots \quad D) \ldots

Output your final answer after 'answer\_choice':''
\end{quote}
\end{tcolorbox}
The system then generates a reasoning chain and ultimately appends the chosen answer. In our experiments, we parse out that final selection to compare with the correct label. More details can be found in our code.

\subsection{Additional Notes and Practical Considerations}
\label{sec:additional}

\paragraph{Ethical, Privacy, and Data Protection Considerations.} 
A critical concern when handling medical questions is avoiding patient data exposure. Directly inputting raw patient information into language models for retrieval risks potential data leakage and privacy violations. Our query rewriting approach addresses this by extracting only disease-related keywords and conceptual terms from the original question, rather than processing sensitive patient details. When retrieving web snippets, we further minimize risks by using only short search engine excerpts (titles and summaries) rather than full documents. This dual protection - \textit{question sanitization} through rewriting and \textit{restricted snippet usage} - prevents sensitive health information from entering the model's processing pipeline while avoiding unintended use of copyrighted materials.

\section{Case Study}
\onecolumn
\begin{tcolorbox}[breakable, title={Case Study 2: }, colback=white, colframe=black, colbacktitle=white, coltitle=black, fonttitle=\bfseries, rounded corners]
\small
\textbf{Question:}
A healthy 23-year-old male is undergoing an exercise stress test as part of his physiology class. If blood were to be sampled at different locations before and after the stress test, which area of the body would contain the lowest oxygen content at both time points?
\vspace{5pt} %

\textbf{Choices:} 
[A.] Inferior vena cava, [B.] Coronary sinus, [C.] IV Pulmonary artery, [D.] Pulmonary vein

\vspace{2pt} % 添加额外的空白间隔

\textbf{Ground Truth: B (Coronary sinus)}
\vspace{2pt} %

\textbf{Synthetic Queries:} 

[1] ‘anatomical variation oxygen levels blood sampled', \textcolor{gray}{[2] 'Oxygen content in different body compartments during exercise'}, \textcolor{gray}{[3] 'Oxygen levels during exercise in pulmonary circulation'}, \textcolor{gray}{[4] 'Oxygen content in various body veins during exercise'}, \textcolor{gray}{5] 'Oxygen content blood different body locations'}, [6] 'blood oxygen content vs venous return areas', \textcolor{gray}{[7] 'Blood oxygen content after exercise physiology'}, [8] 'oxygen content gradient in coronary circulation'.
\vspace{2pt} %

\textbf{Retrieved Fact:} 

[1] Oxygen Saturation - StatPearls - NCBI Bookshelf - In the case of hypoxemia, a physician should always consider whether an arterial blood sample would provide a more accurate measure of oxygen saturation than ...
Pattern Analysis of Oxygen Saturation Variability in Healthy Individuals - Pulse oximetry is routinely used for monitoring patients' oxygen saturation levels with little regard to the variability of this physiological ...
ABG, VBG, and pulse oximetry - EMCrit Project - The oxygen extraction of the hand isn't very high, so a \hlgreen{venous sample will often have only a slightly lower oxygen level than the arterial blood} ...

[6] Venous Oxygen Saturation - StatPearls - NCBI Bookshelf - \hlgreen{Venous oxygen saturation (SvO2) measures the blood's oxygen content} returning to the right side of the heart after perfusing the entire body.
\hlgreen{Arteriovenous Oxygen Difference} - an overview | ScienceDirect Topics - The arteriovenous oxygen difference is a measure of the amount of oxygen taken up from the blood by the tissues.
Relating oxygen partial pressure, saturation and content - Consequently, the blood “reserve” of oxygen is diminished and the venous oxygen content, saturation and partial pressure are all less than normal.

[8] Physiology, Alveolar to Arterial Oxygen Gradient - StatPearls - NCBI - The A-a gradient, or the alveolar-arterial gradient, measures the difference between the oxygen concentration in the alveoli and arterial system.
Regulation of Coronary Blood Flow - PMC - PubMed Central - In a typical adult heart, baseline/resting left coronary blood flow typically ranges between ~0.5 and 1.0 mL/min/g (270, 331, 918, 919). In contrast, right ...
Coronary Artery Pressure - an overview | ScienceDirect Topics - \hlgreen{The high extraction reduces the oxygen content from 195 ml/litre in arterial blood to only 50–70 ml/litre in coronary sinus blood}.

\vspace{2pt}

\textbf{Answer Before Retrieval:} \\
To determine which area of the body would contain the lowest oxygen content at both time points before and after the stress test, we need to consider the path of oxygen in the body during exercise. The oxygen content in the blood is typically highest in the pulmonary artery, as it is the point of entry of oxygenated blood from the lungs. As the blood travels through the body, the oxygen content decreases due to the extraction of oxygen by the tissues. The coronary sinus, which is the collection of blood from the myocardium, would have a lower oxygen content than the pulmonary artery because the myocardium extracts oxygen from the blood. The inferior vena cava, which carries deoxygenated blood from the body back to the heart, would have an even lower oxygen content than the coronary sinus. The pulmonary vein, which carries oxygenated blood from the heart to the lungs, would have a higher oxygen content than the pulmonary artery. Therefore, the area of the body that would contain the lowest oxygen content at both time points before and after the stress test is the \hlsalmon{inferior vena cava}.
\vspace{2pt} 

\textbf{Answer after Retrieval:} \\
To determine which area of the body would contain the lowest oxygen content at both time points, we need to consider the oxygen content gradient in different parts of the body. 

1. First, we need to understand that the oxygen content in the blood decreases as it moves from the arterial system to the venous system due to the extraction of oxygen by the tissues. This is known as the \hlgreen{arteriovenous oxygen difference}.

2. Next, we need to consider the oxygen content in different venous return areas. \hlgreen{The venous oxygen saturation (SvO2) measures the blood's oxygen content} returning to the right side of the heart after perfusing the entire body. However, the oxygen content in the venous blood can vary depending on the location of the sampling site.

3. We are given information about the oxygen content in the coronary sinus, which is the venous blood returning from the heart muscle itself. The oxygen content in the coronary sinus is significantly lower than in other venous return areas, \hlgreen{with a value of only 50-70 ml/litre compared to 195 ml/litre in arterial blood}.

4. Considering the information provided, we can conclude that the area of the body that would contain the \hlgreen{lowest oxygen content at both time points is the coronary sinus}, as it has the lowest oxygen content among the given options.

5. Therefore, the correct answer is the area that has the lowest oxygen content, which is the \hlsalmon{coronary sinus}.

\end{tcolorbox}

\newpage

\begin{tcolorbox}[breakable, title={Case Study 3:}, colback=white, colframe=black, colbacktitle=white, coltitle=black, fonttitle=\bfseries, rounded corners]
\small
\textbf{Question:}
A 37-year-old man with no significant past medical history is rear-ended in a motor vehicle accident. He reported significant neck pain to emergency responders, but otherwise denies weakness, numbness or tingling in his extremities. His vitals on presentation to the ED are HR 90, BP 140/80, RR 20, SpO2 98\%. What is the most appropriate next step upon presentation to the emergency room?
\vspace{5pt} %

\textbf{Choices:} 
[A.] Lateral cervical film, [B.] Cervical immobilization, [C.] IV methylprednisolone, [D.] Observation overnight

\vspace{2pt} 

\textbf{Ground Truth: B (Cervical immobilization)}
\vspace{2pt} %

\textbf{Retrieved Queries:} 

[1] ‘Acute trauma cervical spine care guidelines', \textcolor{gray}{[2] 'Cervical spine injury ED management guidelines'}, \textcolor{gray}{[3] 'whiplash injury cervical spine trauma protocol'}, \textcolor{gray}{[4] 'Emergency management after motor vehicle accidents'}, \textcolor{gray}{[5] 'trauma c-spine emergency department protocols'}, [6] 'Emergency treatment motor vehicle accident spinal injury', \textcolor{gray}{[7] 'Motor vehicle cervical spine emergency evaluation guidelines'}, [8] 'Emergency cervical spine injury management'
\vspace{2pt} %

\textbf{Retrieved Fact:} 

[1] Primary survey and \hlgreen{immobilization of the cervical spine}\footnotemark[1] in rigid cervical collar with full spine precautions, which includes placing the patient on a flat surface and log-rolling the patient when needed.

[6] Treating spinal cord injuries. At an accident scene, if SCI is suspected, emergency personnel will \hlgreen{place a rigid collar around the neck} and carefully place the person on a backboard to \hlgreen{prevent further damage} to the spinal cord. They may use sedatives to relax the person and prevent movement.

[8] If you suspect someone has a spinal injury: Call 911 or emergency medical help. Keep the person still. Place heavy towels or rolled sheets on both sides of the neck or \hlgreen{hold the head and neck to prevent movement}. Avoid moving the head or neck.
\vspace{2pt}

\textbf{Answer Before Retrieval:} \\
The patient is a 37-year-old man with no significant past medical history who was rear-ended in a motor vehicle accident. He reports significant neck pain but denies weakness, numbness, or tingling in his extremities. The initial assessment of his vitals is within normal limits. Given the mechanism of injury and the patient's symptoms, the primary concern is for potential cervical spine injury. The Canadian C-Spine Rule is a widely used clinical decision rule for determining the need for cervical spine imaging in patients with acute trauma. According to the rule, patients who meet any of the following criteria should undergo imaging: age 65 or older, dangerous mechanism of injury, paresthesias in the extremities, or a high-risk injury (e.g., fall from a height, axial load to the head or neck). The patient does not meet any of these criteria, but it is still important to consider the possibility of a cervical spine injury. The most appropriate next step is to perform a careful physical examination, including a thorough assessment of the patient's neck and spine. This should include a log roll examination to assess for any signs of cervical spine injury. If the examination is normal, the patient can be considered for observation. However, if there are any signs of cervical spine injury, such as tenderness, deformity, or neurological deficits, imaging should be obtained. In this case, the patient's symptoms and examination are normal, so the most appropriate next step is to \hlsalmon{observe him overnight} and repeat the examination in the morning.
% \vspace{2pt} 

\textbf{Answer after Retrieval:} \\
The patient is suspected to have a cervical spine injury due to the motor vehicle accident. The most appropriate initial step is to prevent further damage to the spinal cord. According to the provided documents, emergency personnel should \hlgreen{place a rigid collar around the neck} and carefully place the person on a backboard to \hlgreen{prevent further damage} to the spinal cord. This is also supported by the acute trauma cervical spine care guidelines, which emphasize the importance of \hlgreen{immobilization of the cervical spine} in a rigid cervical collar with full spine precautions. Therefore, the next step upon presentation to the emergency room should be to \hlsalmon{ensure the patient's cervical spine is immobilized}.

\end{tcolorbox}

\twocolumn

\end{document}